%% file: main.tex
\documentclass[10pt,twocolumn,letterpaper]{article}

\usepackage[pagenumbers]{cvpr} 

\input{preamble}
\definecolor{cvprblue}{rgb}{0.21,0.49,0.74}
\usepackage[pagebackref,breaklinks,colorlinks,allcolors=cvprblue]{hyperref}
\usepackage{graphicx}
\usepackage{amsmath} 
\usepackage{amssymb} 
\usepackage[ruled,vlined]{algorithm2e} 
\usepackage{multirow} 
\usepackage{pifont}
\usepackage{tabularx}
\usepackage[utf8]{inputenc}
\usepackage{cuted}
\usepackage{capt-of}
\usepackage{nth}
\def\paperID{*****} 
\def\confName{CVPR}
\def\confYear{2026}

\title{TurboTalk: Progressive Distillation for One-Step Audio-Driven \\Talking Avatar Generation}

\author{Xiangyu Liu$^{1,2,3}$ \qquad Feng Gao$^{3}$ \qquad Xiaomei Zhang$^{1,2}$ \qquad \\
Yong Zhang$^{3}$\footnotemark[2] \qquad  Xiaoming Wei$^{3}$  \qquad Zhen Lei$^{1,2,4,5}$ \qquad Xiangyu Zhu$^{1,2}$ \footnotemark[2]  \\
$^1$MAIS, Institute of Automation, Chinese Academy of Sciences
\\
$^2$School of Artificial Intelligence, University of Chinese Academy of Sciences\\
$^3$Meituan \qquad
$^4$The Hong Kong Polytechnic University\qquad $^5$CAIR, HKISI, CAS  
}

\begin{document}
\maketitle
\renewcommand{\thefootnote}{\fnsymbol{footnote}}
\footnotetext[2]{Corresponding Authors.}
\begin{strip}
\vspace{-2cm}
\centering
\includegraphics[width=1.0\linewidth]{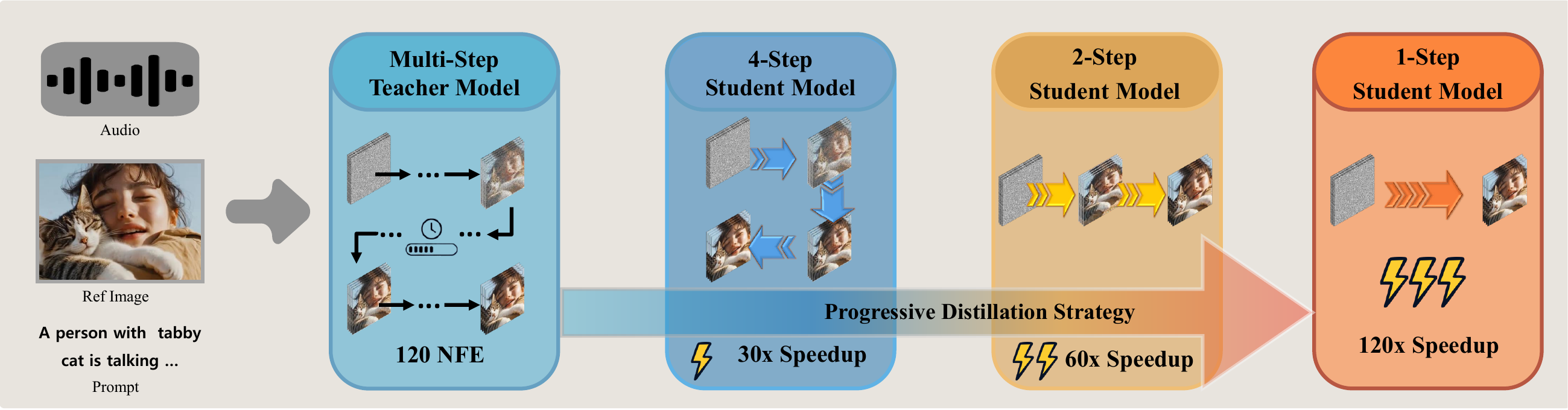}
\setlength{\abovecaptionskip}{0cm} 
\captionof{figure}{We propose TurboTalk, a progressive distillation framework for audio-driven talking avatar generation.}
\label{fig:first}
\end{strip}
\input{sec/0_abs}    
\input{sec/1_intro}
\input{sec/2_related_work}

\input{sec/3_method}
\input{sec/4_exp}

\input{sec/5_conclusion}
{
    \small
    \bibliographystyle{ieeenat_fullname}
    \bibliography{main}
}


\end{document}

%% file: sec/0_abs.tex
\begin{abstract}
Existing audio-driven video digital human generation models rely on multi-step denoising, resulting in substantial computational overhead that severely limits their deployment in real-world settings. While one-step distillation approaches can significantly accelerate inference, they often suffer from training instability. To address this challenge, we propose TurboTalk, a two-stage progressive distillation framework that effectively compresses a multi-step audio-driven video diffusion model into a single-step generator. We first adopt Distribution Matching Distillation to obtain a strong and stable 4-step student, and then progressively reduce the denoising steps from 4 to 1 through adversarial distillation. To ensure stable training under extreme step reduction, we introduce a progressive timestep sampling strategy and a self-compare adversarial objective that provides an intermediate adversarial reference that stabilizes progressive distillation. Our method achieve single-step generation of video talking avatar, boosting inference speed by 120 times while maintaining high generation quality.
\end{abstract}

%% file: sec/1_intro.tex
\section{Introduction}

Audio-driven talking avatar video generation aims to synthesize realistic and temporally coherent human-centric videos directly from speech signals, and has become a foundational technology for interactive digital communication. It underpins a wide range of applications, including virtual anchoring, telepresence, and embodied human–computer interaction, where both visual fidelity and audio–motion synchronization are critical.

Recent advances~\cite{wan2025,liu2024sorareviewbackgroundtechnology,kong2024hunyuanvideo,ho2022videodiffusionmodel} in diffusion-based video generation have significantly improved the visual quality and expressiveness of audio-driven avatars, enabling detailed facial dynamics and natural motion. However, these gains come at the cost of expensive multi-step inference, resulting in high latency and computational overhead that severely limit real-time and streaming deployment. Bridging the gap between high-fidelity biometric animation and ultra-low-latency generation remains a central challenge for practical audio-driven digital human systems.

To overcome the high inference cost caused by the long denoising trajectories of diffusion models, model distillation has emerged as one of the most effective acceleration strategies. By distilling a multi-step diffusion model into a few-step or even single-step student, prior works~\cite{yin2025slow,huang2025self,cui2025self,chen2024diffusion} significantly improve generation efficiency while largely preserving the visual quality of the teacher model. However, existing talking avatar video generation approaches~\cite{shen2025soulxflashtalktechnicalreport,huang2025liveavatarstreamingrealtime,xu2024hallo,yang2025infinitetalk} primarily focus on distilling multi-step diffusion models into four-step students by aligning the prediction distributions of the student and teacher models. Despite this reduction, four-step video diffusion models remain computationally expensive in practice due to the large model scale typically required for high-fidelity video synthesis.

Further reducing the denoising steps to one or two poses substantial challenges. When distribution matching distillation is directly applied under such extremely few-step settings, the discrepancy between the student and teacher distributions becomes excessively large, often leading to unstable training dynamics and severe degradation in generation quality. Recent studies~\cite{lin2025autoregressive,lin2025diffusion,cheng2025pose} attempt to address this issue by leveraging adversarial distillation to directly obtain one-step video generators. Nevertheless, such approaches are notoriously difficult to train, as the large gap between multi-step and single-step outputs causes the discriminator to converge prematurely, resulting in uninformative or vanishing gradient signals during early training stages.

To address these challenges, we propose TurboTalk, a two-stage progressive distillation framework that enables stable and effective distillation of multi-step talking human video diffusion models into a single-step generator. In the first stage, we adopt Distribution Matching Distillation to distill a multi-step teacher into a four-step student model. Specifically, an auxiliary critic network is trained to estimate the score function of the student’s generated distribution, and the student model is optimized by minimizing the Kullback–Leibler divergence between the student and teacher predictive distributions. This stage aims to obtain a strong and stable few-step baseline while preserving the generation quality of the original teacher model.

Building upon the four-step model, we further introduce a progressive adversarial distillation strategy that gradually compresses the denoising steps from four to three, two, and ultimately one. Our approach is centered around three key components.
(a) \textbf{Progressive Step Reduction.} Instead of aggressively distilling a multi-step model directly into a single-step generator, we design a staged training scheme in which each phase reduces only one denoising step. This controlled reduction ensures that the quality gap between consecutive stages remains manageable, allowing the discriminator to provide meaningful gradients and preventing early saturation caused by overly distinct real and generated samples.
(b) \textbf{Dynamic Timestep Sampling.} During the warm-up phase of each stage, we randomly perturb the target timestep rather than fixing it to the exact timestep corresponding to the reduced step count. This strategy encourages the model to learn denoising behaviors over a broader range of timesteps, improving robustness and effectively alleviating training instability induced by abrupt step reduction.
(c) \textbf{Self-Adversarial Regularization.} To prevent the distilled student from deviating excessively from the original multi-step generation distribution, we introduce a self-compare mechanism. In addition to adversarial training against real data, the student is also adversarially aligned with high-quality samples generated by a discriminator equipped with four-step denoising capability. This intermediate supervision signal, lying between real data and the student’s current outputs, significantly reduces optimization difficulty and enables a smooth transition in generation quality across distillation stages.

Through these designs, our method achieves progressive knowledge transfer from multi-step diffusion models to a single-step generator, substantially improving inference efficiency while maintaining high visual fidelity and accurate audio–visual synchronization in talking human video generation.

The main contributions can be summarized as follows:
\begin{itemize}
\item We propose a two-stage progressive distillation framework that enables stable single-step talking human video generation from large multi-step diffusion models.

\item We introduce a progressive adversarial distillation strategy with dynamic timestep sampling and self-adversarial regularization, effectively mitigating training instability and quality collapse under one-step generation.

\item Extensive experiments demonstrate that our method significantly improves inference efficiency while preserving visual quality and audio–visual synchronization.
\end{itemize}

%% file: sec/2_related_work.tex
\section{Related Work}
\textbf{Audio-driven Avatar Video Generation.} Audio-driven avatar video generation aims to synthesize realistic human animations conditioned on speech signals, requiring accurate lip synchronization, identity preservation, and coherent motion dynamics. Wav2Lip~\cite{prajwal2020lip} and SadTalker~\cite{zhang2023sadtalker} typically adopt two-stage pipelines, where audio signals are first mapped to intermediate motion representations such as 3DMM~\cite{tran2018nonlinear} or FLAME~\cite{Flame} parameters, followed by GAN-based rendering to generate talking videos. With the success of diffusion models, recent works~\cite{tian2024emo,wei2024aniportrait,xu2024hallo,wang2025fantasytalking} shift toward end-to-end audio-to-video synthesis, directly integrating audio cues into a single diffusion framework, and DiT-based architectures have shown strong performance for high-fidelity talking head generation. In parallel, recent efforts explore multi-person and conversational avatar generation by binding multiple speakers to multiple audio streams ~\cite{kong2025let}. Additionally, InfiniteTalk~\cite{yang2025infinitetalk} investigates long-form and cinematic audio-driven video generation, including sparse-frame video dubbing for long sequences and Wan2.2-S2V~\cite{gao2025wans2vaudiodrivencinematicvideo} investigates cinematic-level character animation with complex motion and camera dynamics.

\textbf{Distribution Matching Distillation.} Distribution Matching Distillation~\cite{yin2024one} aligns the student's generation distribution with that of a pretrained diffusion teacher at given noise levels, enabling effective few-step acceleration while preserving generative priors. Originally developed for large-scale image diffusion models, DMD has inspired several extensions, including adversarial variants \cite{yin2024improved} and trajectory-aware formulations~\cite{luo2025learning}, to improve alignment quality and support multi-step distillation. In the video domain, DMD has been widely adopted for few-step and streaming video generation, where methods such as CausVID~\cite{yin2025slow}, Self-Forcing~\cite{huang2025self}, and LongLive\cite{yang2025longlive} leverage DMD-based distillation to accelerate sampling while maintaining long-form video quality.

\begin{figure*}[t]
  \centering
  \includegraphics[width=1.0\textwidth]{pipeline.pdf}
  \caption{
 Overview of TurboTalk. We first distill a multi-step diffusion teacher into a 4-step student via distribution matching, and then progressively reduce the denoising steps to a single step through adversarial distillation with step reduction, dynamic timestep sampling and self-compare regularization.
  }
  \label{fig:pipeline}
\end{figure*}

\textbf{Adversarial Distillation.} Adversarial distillation was first shown to be effective in the image domain. Early methods such as ADD~\cite{sauer2024adversarial} and UFOGen~\cite{xu2024ufogen} align the student with real data by matching the final generation outputs, while later works, including SDXL-Lightning~\cite{lin2024sdxl} and LADD~\cite{sauer2024fast}, further improve distillation quality by enforcing consistency over intermediate denoising trajectories between the student and teacher models.

Recently, adversarial distillation has been extended to video generation, particularly for extremely few-step and one-step settings. APT~\cite{lin2025diffusion} introduces adversarial fine-tuning against real video data to enable one-step video generation with improved stability and quality. SAD~\cite{yang2025ASD} proposes an Adversarial Self-Distillation framework that aligns different denoising steps within the student model, providing smoother supervision for one-step and two-step video synthesis. POSE~\cite{cheng2025pose} further explores phased adversarial distillation for large-scale video diffusion models, combining stability priming and self-adversarial equilibrium to achieve high-quality single-step video generation. In addition, APT2\cite{lin2025autoregressive} investigates adversarial training for autoregressive video generation, enabling real-time and interactive video synthesis with one-step neural function evaluations.

%% file: sec/3_method.tex
\section{Method}

We present \textbf{TurboTalk}, a two-stage progressive distillation framework
that compresses a multi-step audio-driven talking avatar diffusion model into an
efficient single-step generator. As illustrated in
Fig.~\ref{fig:pipeline}, our approach first applies
Distribution Matching Distillation (Sec.~\ref{Preliminaries}) to transfer generator from a multi-step teacher to a four-step student.
Building upon this intermediate model, we further perform
Progressive Adversarial Distillation (Sec.~\ref{PAD}) to gradually reduce the
number of denoising steps from four to one.
To stabilize training under such extreme step reduction, we introduce
Dynamic Timestep Sampling (Sec.~\ref{DTM}) to smooth stage transitions, together
with a Self-Compare Regularization mechanism (Sec.~\ref{Self-Compare}) that constrains
the student to remain close to a high-quality multi-step reference.

\subsection{Preliminaries: 4-Step Audio-Driven Video Generation}
\label{Preliminaries}
\textbf{Audio-driven Video Generation.} We incorporate audio signals into video generation by augmenting an image-to-video diffusion backbone with audio cross-attention. Specifically, an additional audio cross-attention layer is inserted after the text cross-attention in each DiT block, enabling frame-wise modulation of visual dynamics by speech.

Audio features are extracted using a pretrained Wav2Vec2 encoder. Since facial motion depends on both past and future speech, we construct a temporal audio context for each frame by concatenating neighboring audio features:
\begin{equation}
  \mathbf{a}_i = \mathrm{Concat}(\mathbf{a}_{i - \lfloor k/2 \rfloor}, \ldots, \mathbf{a}_{i + \lfloor k/2 \rfloor}),
\end{equation}
where $k$ denotes the context length.

Due to temporal compression in video latents, the audio sequence is longer than the video sequence. To bridge this gap, we introduce a lightweight audio adapter that compresses audio embeddings along the temporal dimension via downsampling and MLP encoding:
\begin{equation}
  \mathbf{c}_a = \mathrm{MLP}(\mathrm{Concat}(\mathrm{MLP}(\mathbf{a}_1),\, \mathrm{MLP}(\mathrm{Down}(\mathbf{a}_{2:l})))).
\end{equation}

\textbf{Matching Distillation for 4-Step Generation.}
Multi-step denoising and classifier-free guidance introduce substantial computational overhead in audio-driven video diffusion models. To address this issue, we adopt Distribution Matching Distillation to compress a multi-step teacher into a 4-step student generator, enabling efficient inference while preserving conditional generation quality.

Specifically, DMD minimizes the distributional discrepancy between the teacher and student at each noise level 
$t$ by optimizing the reverse Kullback-Leibler divergence. The resulting objective admits a tractable score-based gradient:
\begin{equation}
  \nabla_\phi \mathcal{L}_{\mathrm{DMD}} = -\mathbb{E}_{t, \epsilon}
\left[ \left( s_r(\Psi(G_\phi)) - s_f(\Psi(G_\phi)) \right) \frac{\partial G_\phi}{\partial \phi} \right]
\end{equation}
where $s_{\mathrm{r}}$ and $s_{\mathrm{f}}$ are the score functions estimated by the teacher model and the trainable critic. $G_\phi$ denotes the 4-step student generator
 and $\Psi$ represents the forward diffusion process. 

\subsection{Progressive Adversarial Distillation}
\label{PAD}

Although the four-step model significantly reduces inference cost, further compression is required to enable real-time generation. Directly distilling a four-step model into a single-step generator often leads to unstable training, as the large discrepancy between generated samples causes the discriminator to converge prematurely and provide ineffective gradients. To address this challenge, we propose progressive adversarial distillation.

\textbf{Step Reduction.}
We perform progressive step reduction over three phases, where each phase decreases the number of denoising steps by one to ensure stable adversarial training.
Let $\mathcal{T}^{(k)} = \{ t_0^{(k)}, \ldots, t_{N_k-1}^{(k)} \}$ denote the timestep schedule of phase $k$, with $N_k \in \{3, 2, 1\}$ and
$\mathcal{T}^{(1)}=\{1.0, 0.75, 0.5\}$,
$\mathcal{T}^{(2)}=\{1.0, 0.75\}$,
$\mathcal{T}^{(3)}=\{1.0\}$.

For each phase $k$, we generate samples by executing the full
$N_k$-step denoising trajectory, starting from Gaussian noise
$\mathbf{z}_0 \sim \mathcal{N}(\mathbf{0}, \mathbf{I})$.
However, during backpropagation, gradients are intentionally
restricted to the final denoising step only.
Formally, the denoising update is written as
\begin{equation}
\mathbf{z}_{i+1}
=
\mathbf{z}_i - \tilde{\mathbf{v}}_\theta(\mathbf{z}_i, t_i)\,(t_i - t_{i+1}),
\end{equation}
where
\begin{equation}
\tilde{\mathbf{v}}_\theta(\mathbf{z}_i, t_i)
=
\begin{cases}
\texttt{stop\_grad}(\mathbf{v}_\theta(\mathbf{z}_i, t_i)), & i < N_k-1, \\
\mathbf{v}_\theta(\mathbf{z}_i, t_i), & i = N_k-1.
\end{cases}
\end{equation}

This design ensures that intermediate denoising steps are treated
as fixed transformations, while optimization focuses exclusively
on the final step that directly determines the output sample
$\hat{\mathbf{x}}_0 = \mathbf{z}_{N_k}$.

As a result, it significantly reduces memory consumption by avoiding
the storage of intermediate activations, while concentrating
optimization on the most critical denoising step.
Moreover, combined with progressive step reduction, this strategy
keeps the quality gap between consecutive phases small, preventing
discriminator saturation and enabling stable adversarial training
under extreme step reduction.





\textbf{Adversarial Training with R3GAN.}
We adopt R3GAN~\cite{huang2024gan} as the adversarial objective, which formulates discrimination in a
relativistic manner by comparing the relative realism between real and generated samples,
leading to more stable training when the two distributions become close.

\begin{equation}
\begin{aligned}
  \mathcal{L}_D
  &= -\mathbb{E}\!\left[
      \log \sigma\!\left(
      D(\mathbf{x}_{\mathrm{real}}) - D(\hat{\mathbf{x}}_0)
      \right)
     \right], \\
  \mathcal{L}^{\mathrm{real}}_G
  &= -\mathbb{E}\!\left[
      \log \sigma\!\left(
      D(\hat{\mathbf{x}}_0) - D(\mathbf{x}_{\mathrm{real}})
      \right)
     \right],
\end{aligned}
\end{equation}
where $D(\cdot)$ denotes the discriminator, $\sigma(\cdot)$ is the sigmoid function,
$\mathbf{x}_{\mathrm{real}}$ is a real video sample,
and $\hat{\mathbf{x}}_0$ is the generated output of the student model.



To mitigate discriminator overfitting and improve training stability, we extend the standard R1/R2
regularization scheme to all discriminator inputs involved in progressive
distillation. In addition to real samples and student-generated samples,
we also regularize the discriminator response on the self-compare reference
samples produced by the four-step model in Sec~\ref{Self-Compare}.

Specifically, we define the following three regularization terms:
\begin{equation}
\begin{aligned}
  \mathcal{L}_{\mathrm{R1}}
  &= \mathbb{E}\!\left[
      \left\lVert
      D(\mathbf{x}_{\mathrm{real}} + \sigma_r \boldsymbol{\epsilon})
      - D(\mathbf{x}_{\mathrm{real}})
      \right\rVert^2
     \right], \\
  \mathcal{L}_{\mathrm{R2}}
  &= \mathbb{E}\!\left[
      \left\lVert
      D(\hat{\mathbf{x}}_0 + \sigma_r \boldsymbol{\epsilon})
      - D(\hat{\mathbf{x}}_0)
      \right\rVert^2
     \right], \\
  \mathcal{L}_{\mathrm{R3}}
  &= \mathbb{E}\!\left[
      \left\lVert
      D(\hat{\mathbf{x}}_0^{4\text{-step}} + \sigma_r \boldsymbol{\epsilon})
      - D(\hat{\mathbf{x}}_0^{4\text{-step}})
      \right\rVert^2
     \right],
\end{aligned}
\end{equation}
where $\boldsymbol{\epsilon} \sim \mathcal{N}(\mathbf{0}, \mathbf{I})$ is Gaussian
noise and $\sigma_r$ controls the perturbation magnitude.
Here, $\hat{\mathbf{x}}_0$ denotes the output of the $n$-step student model,
and $\hat{\mathbf{x}}_0^{4\text{-step}}$ is the self-compare reference generated
by the four-step model.

The final discriminator objective is given by
\begin{equation}
  \mathcal{L}_D^{\mathrm{total}}
  =
  \mathcal{L}_D
  +
  \frac{\gamma}{3}
  \left(
  \mathcal{L}_{\mathrm{R1}}
  +
  \mathcal{L}_{\mathrm{R2}}
  +
  \mathcal{L}_{\mathrm{R3}}
  \right),
\end{equation}
where $\gamma$ controls the overall strength of discriminator regularization.

\subsection{Dynamic Timestep Sampling}
\label{DTM}

At transitions between progressive distillation stages, the student must adapt
to a reduced number of denoising steps. Directly switching to a fixed timestep
schedule often results in unstable optimization due to the abrupt increase in
denoising difficulty. To address this issue, we introduce \emph{dynamic timestep
sampling}, which serves as a curriculum over timestep gaps.

For each distillation stage $k \in \{1,2,3\}$, let
$\mathcal{T}^{(k)} = \{t_0^{(k)}, \ldots, t_{N_k-1}^{(k)}\}$ denote the target
timestep schedule, and let $W_k$ denote the number of warm-up optimization steps.
We use $s$ to denote the current training iteration within stage $k$.
During the warm-up phase ($s < W_k$), the final timestep in $\mathcal{T}^{(k)}$
is randomly perturbed as
\begin{equation}
t_{N_k-1}^{(k)} \sim \mathcal{U}\!\left(
t_{N_{k-1}-1}^{(k-1)},
t_{N_k-1}^{(k)}
\right),
\end{equation}
while all other timesteps remain fixed.
After warm-up ($s \geq W_k$), the schedule is fixed to the target
$\mathcal{T}^{(k)}$.

This strategy gradually bridges the denoising difficulty between consecutive
stages, enabling smoother adaptation to larger timestep gaps and significantly
improving training stability during extreme step reduction.

\subsection{Self-Compare Regularization}
\label{Self-Compare}
A common issue in adversarial distillation is that the student model may drift
excessively from the generative distribution of the original multi-step model,
resulting in uncontrolled quality degradation. To mitigate this effect, we
propose self-compare regularization, which introduces an intermediate
adversarial target derived from the four-step model.

Specifically, the discriminator is augmented with four-step denoising capability
and is able to generate reference samples
$\hat{\mathbf{x}}_0^{4\text{-step}}$. During training, an $N$-step student generator
($N < 4$) is optimized adversarially not only against real data
$\mathbf{x}_{\text{real}}$, but also against the four-step reference samples:
\begin{equation}
  \mathcal{L}_G^{\mathrm{self}}
  =
  -\mathbb{E}\!\left[
    \log \sigma\!\left(
      D(\hat{\mathbf{x}}_0^{N}) - D(\hat{\mathbf{x}}_0^{4\text{-step}})
    \right)
  \right].
\end{equation}

This design provides an intermediate supervisory signal that lies between real
data and the student’s current outputs. Since four-step samples exhibit higher
visual fidelity than those produced by the $N$-step student, yet remain less
ideal than real data, they offer smoother and more fine-grained guidance during
adversarial optimization.

The final adversarial objective for the generator is defined as a weighted
combination of the standard adversarial loss and the self-compare regularization:
\begin{equation}
  \mathcal{L}_G^{\mathrm{adv\text{-}total}}
  =
  (1 - \lambda)\,\mathcal{L}_G^{\mathrm{real}}
  +
  \lambda\,\mathcal{L}_G^{\mathrm{self}},
\end{equation}
where $\lambda$ controls the strength of the self-compare regularization.

\begin{algorithm}[t]
\caption{Progressive Adversarial Distillation}
\label{alg:turbotalk}
\KwIn{4-step model $G_{\phi_0}$, discriminator $D_\psi$, data $\mathcal{D}$}
\KwOut{Single-step generator $G_\phi$}

Initialize $G_\phi \leftarrow G_{\phi_0}$; \quad $D_\psi \leftarrow G_{\phi_0}$

$\mathcal{T}^{(1)}\!=\!\{1.0, 0.75, 0.5\}$, $\mathcal{T}^{(2)}\!=\!\{1.0, 0.75\}$, $\mathcal{T}^{(3)}\!=\!\{1.0\}$

\For{phase $k = 1, 2, 3$}{
    $N_k \leftarrow |\mathcal{T}^{(k)}|$ 
    
    \For{each training step $s$}{
        $\mathcal{T} \leftarrow \texttt{DynamicSample}(\mathcal{T}^{(k)}, s)$ 
        
        Sample $(\mathbf{x}_{\mathrm{real}}, \mathbf{c}) \sim \mathcal{D}$; \quad $\mathbf{z}_0 \sim \mathcal{N}(\mathbf{0}, \mathbf{I})$
        
        \tcp{$N_k$-step denoising}
        \For{$i = 0$ \KwTo $N_k - 1$}{
            $\tilde{\mathbf{v}}_\theta \leftarrow 
            \begin{cases}
            \texttt{stop\_grad}(\mathbf{v}_\theta(\mathbf{z}_i, t_i, \mathbf{c})), & i < N_k-1 \\
            \mathbf{v}_\theta(\mathbf{z}_i, t_i, \mathbf{c}), & i = N_k-1
            \end{cases}$
            
            $\mathbf{z}_{i+1} \leftarrow \mathbf{z}_i - \tilde{\mathbf{v}}_\theta \cdot (t_i - t_{i+1})$
        }
        $\hat{\mathbf{x}}_0 \leftarrow \mathbf{z}_{N_k}$
        
        \tcp{$4$-step self-compare}
        $\hat{\mathbf{x}}_0^{\text{4-step}} \leftarrow \texttt{Denoise}(D_\psi, \mathbf{z}_0, \mathcal{T}^{(0)}, \mathbf{c})$

        $\mathcal{L}_G \leftarrow \mathcal{L}_G^{\mathrm{adv}} + \lambda \mathcal{L}_G^{\mathrm{self}} $ 
        
        $\mathcal{L}_D \leftarrow \mathcal{L}_D^{\mathrm{R3GAN}} + \gamma \mathcal{L}_{\mathrm{reg}}$ 
        
        Update $G_\phi$, $D_\psi$ with $\nabla \mathcal{L}_G$, $\nabla \mathcal{L}_D$
    }
}
\Return $G_\phi$
\end{algorithm}

The complete training procedure of our progressive adversarial distillation framework, including step reduction, dynamic timestep sampling, and self-compare regularization, is summarized in Algorithm~\ref{alg:turbotalk}.

%% file: sec/4_exp.tex
\begin{table*}[ht]
\centering
    \caption{Quantitative comparison with other audio-driven talking avatar generation methods under different NFE settings.Our method achieves competitive or superior quality even at 1-NFE.$^\ast$ denotes models enhanced with LightX2V LoRA.}
     \label{Table:Quan_exp}
\setlength{\tabcolsep}{4pt} 

\resizebox{\linewidth}{!}{\begin{tabular}{lcccccccccccc}
\hline
\multicolumn{1}{c|}{\multirow{2}{*}{Method}}  & \multicolumn{4}{c|}{HDTF}                                                             & \multicolumn{4}{c|}{CeleHQ}                                                           & \multicolumn{4}{c}{EMTD}                                         \\ \cline{2-13} 
\multicolumn{1}{c|}{}                         & FID↓            & FVD↓             & Sync-C↑        & \multicolumn{1}{c|}{Sync-D↓}        & FID↓            & FVD↓             & Sync-C↑        & \multicolumn{1}{c|}{Sync-D↓}        & FID↓            & FVD↓             & Sync-C↑        & Sync-D↓        \\ \hline
\multicolumn{13}{c}{\textbf{Many NFE}}                                                                                                                                                                                                                                                           \\ \hline
\multicolumn{1}{l|}{Wan2.2-S2V~\cite{gao2025wans2vaudiodrivencinematicvideo}}               & 43.30          & 127.33          & 9.50          & \multicolumn{1}{c|}{8.08}          & 31.71          & 290.34          & 8.45          & \multicolumn{1}{c|}{\textbf{7.21}} & 31.91          & \textbf{358.80} & 8.73          & 7.34          \\
\multicolumn{1}{l|}{InfiniteTalk~\cite{yang2025infinitetalk}}             & \textbf{42.29} & \textbf{110.51} & \textbf{9.81} & \multicolumn{1}{c|}{\textbf{6.96}} & \textbf{30.87} & \textbf{286.36} & \textbf{8.64} & \multicolumn{1}{c|}{7.24}          & \textbf{29.88} & 385.95          & \textbf{8.23} & \textbf{7.29} \\ \hline
\multicolumn{13}{c}{\textbf{4-NFE}}                                                                                                                                                                                                                                                              \\ \hline
\multicolumn{1}{l|}{InfiniteTalk$^\ast$~\cite{yang2025infinitetalk,lightx2v}}             & 47.40          & 167.29          & 8.96          & \multicolumn{1}{c|}{7.00}          & 36.38          & 422.12          & 6.58          & \multicolumn{1}{c|}{7.59}          & 35.31          & 425.14          & 8.32          & \textbf{7.31} \\
\multicolumn{1}{l|}{Live Avatar~\cite{huang2025liveavatarstreamingrealtime}}              & 47.14          & 165.73          & 8.91          & \multicolumn{1}{c|}{7.19}          & 32.30          & 326.10          & 6.11          & \multicolumn{1}{c|}{8.25}          & \textbf{29.35}          & 405.17          & 6.83          & 8.44          \\
\multicolumn{1}{l|}{SoulX-LiveTalk~\cite{shen2025soulxflashtalktechnicalreport}}           & 46.13          & 161.26          & 9.12          & \multicolumn{1}{c|}{7.37}          & 32.15          & 335.94          & 7.70          & \multicolumn{1}{c|}{7.64}          & 29.36 & 422.57          & 8.55          & 7.59          \\
\multicolumn{1}{l|}{\textbf{TurboTalk(Ours)}} & \textbf{43.28} & \textbf{160.79} & \textbf{9.45} & \multicolumn{1}{c|}{\textbf{6.91}} & \textbf{32.14} & \textbf{310.93} & \textbf{8.12} & \multicolumn{1}{c|}{\textbf{7.49}} & 29.41          & \textbf{401.14} & \textbf{8.76} & 7.74          \\ \hline
\multicolumn{13}{c}{\textbf{2-NFE}}                                                                                                                                                                                                                                                              \\ \hline
\multicolumn{1}{l|}{InfiniteTalk$^\ast$~\cite{yang2025infinitetalk,lightx2v}}             & 49.95          & 175.67          & 8.79          & \multicolumn{1}{c|}{6.84}          & 37.73          & 428.03          & 6.70          & \multicolumn{1}{c|}{7.68} & 36.58          & 449.30          & 7.56          & 8.43          \\
\multicolumn{1}{l|}{Live Avatar~\cite{huang2025liveavatarstreamingrealtime}}              & 47.81          & 178.11          & 7.37          & \multicolumn{1}{c|}{8.19}          & 35.45          & 349.11          & 6.35          & \multicolumn{1}{c|}{8.25}          & 32.49          & 412.29          & 6.39          & 8.61          \\
\multicolumn{1}{l|}{SoulX-LiveTalk~\cite{shen2025soulxflashtalktechnicalreport}}           & 49.96          & 175.77          & 9.00          & \multicolumn{1}{c|}{7.20}          & 33.86          & 347.08          & 7.60          & \multicolumn{1}{c|}{8.19}          & 33.83          & 463.02          & 7.86          & 8.41          \\
\multicolumn{1}{l|}{\textbf{TurboTalk(Ours)}} & \textbf{44.18} & \textbf{161.76} & \textbf{9.28} & \multicolumn{1}{c|}{\textbf{6.83}} & \textbf{32.51} & \textbf{314.87} & \textbf{8.07} & \multicolumn{1}{c|}{\textbf{7.63}}          & \textbf{30.12} & \textbf{397.31} & \textbf{8.70} & \textbf{7.77} \\ \hline
\multicolumn{13}{c}{\textbf{1-NFE}}                                                                                                                                                                                                                                                              \\ \hline
\multicolumn{1}{l|}{InfiniteTalk$^\ast$~\cite{yang2025infinitetalk,lightx2v}}             & 58.99          & 233.78          & 7.51          & \multicolumn{1}{c|}{8.00}          & 54.83          & 607.07          & 5.02          & \multicolumn{1}{c|}{8.68}          & 48.15          & 643.50          & 6.35          & 8.64          \\
\multicolumn{1}{l|}{Live Avatar~\cite{huang2025liveavatarstreamingrealtime}}              & 57.00          & 250.85          & 7.35          & \multicolumn{1}{c|}{8.20}          & 38.64          & 394.12          & 6.34          & \multicolumn{1}{c|}{8.51}          & 34.67          & 441.32          & 6.38          & 8.82          \\
\multicolumn{1}{l|}{SoulX-LiveTalk~\cite{shen2025soulxflashtalktechnicalreport}}           & 51.63          & 262.33          & 8.07          & \multicolumn{1}{c|}{7.28}          & 37.36          & 397.55          & 7.60          & \multicolumn{1}{c|}{8.58}          & 35.67          & 477.20          & 6.61          & 8.60          \\
\multicolumn{1}{l|}{\textbf{TurboTalk(Ours)}} & \textbf{45.19} & \textbf{164.84} & \textbf{9.10} & \multicolumn{1}{c|}{\textbf{6.99}} & \textbf{32.15} & \textbf{322.32} & \textbf{8.51} & \multicolumn{1}{c|}{\textbf{7.89}} & \textbf{31.43} & \textbf{404.23} & \textbf{8.50} & \textbf{8.03} \\ \hline
\end{tabular}
}
\end{table*}
\section{Experiments}

\begin{figure*}[ht]
  \centering
  \includegraphics[width=1.0\textwidth]{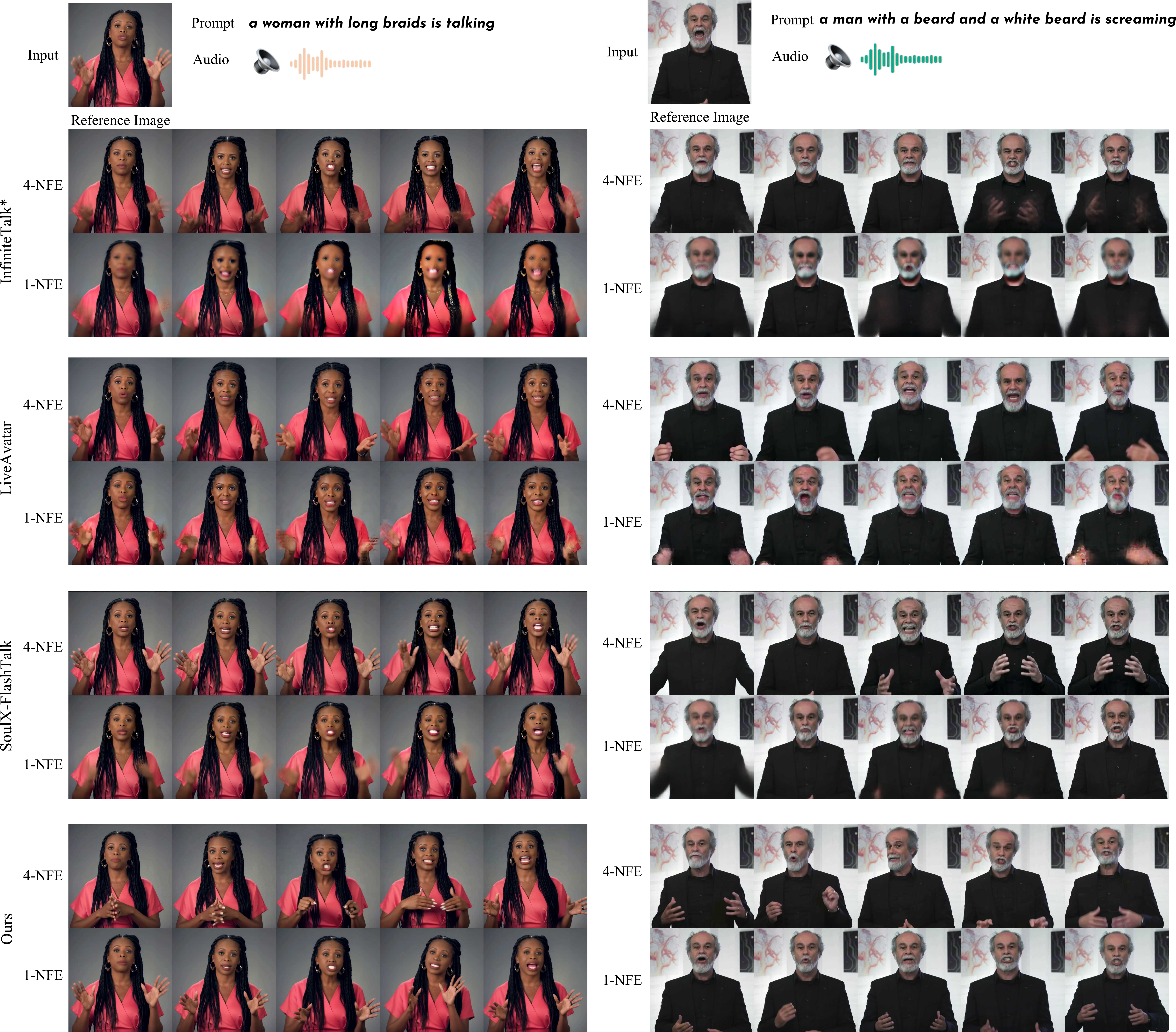}
  \caption{
    Qualitative comparison with other audio-driven talking avatar generation methods under different NFE settings.
    Under the 4-NFE setting, our method produces more dynamic and expressive motions, while under the 1-NFE setting it maintains visual quality comparable to its 4-NFE counterpart.
$^\ast$ denotes models enhanced with LightX2V LoRA.
  }
  \label{fig:qualitive}
\end{figure*}
\subsection{Experimental Settings}
\textbf{Implementation Details.} Our overall model architecture is adapted from InfiniteTalk~\cite{yang2025infinitetalk}. During the DMD distillation stage, both the student generator and the fake score network are initialized from the pretrained InfiniteTalk model. In the progressive adversarial distillation stage, the generator and the discriminator backbone are initialized from the 4-step model obtained after DMD distillation. The discriminator’s classification head follows the design of APT and is zero-initialized at the final layer to stabilize early-stage adversarial training.
For training, the DMD stage is conducted for 2,000 steps using 64 NVIDIA H800 GPUs, while the progressive adversarial distillation stage is trained for 3,000 steps on 32 NVIDIA H800 GPUs. The per-GPU batch size is set to 1. To alleviate memory constraints arising from training multiple large-scale models, all experiments are performed with DeepSpeed ZeRO Stage-3.
We adopt a mixed training strategy in which 50\% of the training iterations include 9-frame context frames to enhance long-range temporal coherence. Each training sample consists of a video chunk of 81 frames, with a mixed resolution setting centered around 720p.
The learning rate for the generator is set to $4\times10^{-7}$, while both the fake score network and the discriminator use a learning rate of $2\times10^{-6}$. In progressive adversarial distillation, the warm-up duration for dynamic timestep sampling is 500 steps. For adversarial regularization, R1 and R2 penalties are applied with  $\sigma_r=0.1$ and $\gamma=100$, and the $\lambda$ for self-compare regularization is set to $0.5$.

\textbf{Datasets.} We collect a large-scale video dataset of approximately 2,000 hours, which is used consistently across all training stages. The dataset primarily consists of videos featuring the face or full body of a single talking person. In addition, we collect an auxiliary dataset of around 200K video clips containing multiple events and diverse human–object or human–environment interactions, with an average clip duration of approximately 10 seconds.
For evaluation, we construct three types of test sets: (i) a talking head dataset, (ii) a talking body dataset, and (iii) a dual-human talking body dataset involving interactive scenarios. For talking head evaluation, we use two publicly available benchmarks, HDTF~\cite{zhang2021flow} and CelebV-HQ~\cite{zhu2022celebv}. For talking body evaluation, we adopt the EMTD~\cite{meng2025echomimicv2} dataset. For each dataset, we randomly sample 40 videos to conduct audio-driven talking avatar generation experiments. During evaluation, all methods generate videos at a unified resolution of the condition image to ensure fair comparison. Our ablation experiments were all conducted on the EMTD dataset, while keeping the same number of training steps.

\textbf{Evaluation Metrics.} We evaluate all methods using commonly adopted quantitative metrics. Fréchet Inception Distance (FID) [44] and Fréchet Video Distance (FVD) [45] are employed to assess the visual quality of the generated images and videos, respectively. Expression-FID (E-FID) is used to measure the expressiveness of facial motions in the generated videos. In addition, Sync-C and Sync-D [46] are adopted to evaluate the synchronization between the input audio and the generated lip movements.

\subsection{Main Result}

\textbf{Quantitative Results.} We compare TurboTalk with recent state-of-the-art audio-driven avatar video generation and acceleration methods. For multi-step diffusion models, we select InfiniteTalk~\cite{yang2025infinitetalk} and Wan2.2-S2V~\cite{gao2025wans2vaudiodrivencinematicvideo}, where InfiniteTalk also serves as our primary baseline due to its architectural similarity to our framework. For few-step models, we include a 4-step accelerated version of InfiniteTalk enhanced with the LightX2V~\cite{lightx2v} LoRA, as well as LiveAvatar~\cite{huang2025liveavatarstreamingrealtime} and SoulX-FlashTalk~\cite{shen2025soulxflashtalktechnicalreport}, which are widely adopted in the community. These few-step baselines are originally trained with 4-NFE(Number of Function Evaluation). Since there are currently no existing audio-driven avatar video generation models explicitly designed for 2-NFE or 1-NFE inference, we directly evaluate these 4-NFE models under 2-NFE and 1-NFE settings and compare them with our method. All selected methods are derived from the Wan models~\cite{wan2025} and share similar architectural scales, with model sizes of approximately 14B parameters. Under this setting, inference speed can be fairly and directly compared using the number of the NFE.

Table \ref{Table:Quan_exp} reports the quantitative comparison between our method and state-of-the-art audio-driven talking avatar generation approaches. For many-NFE models, the use of classifier-free guidance  incurs substantial computational overhead: WAN2.2-S2V and InfiniteTalk require approximately 80 and 120 NFE, respectively, to generate a single video chunk. In contrast, our method achieves high-quality talking avatar generation with as few as 1 NFE, corresponding to a 120× inference speedup over the InfiniteTalk baseline.

Under the 4-NFE setting, our model consistently outperforms other distillation-based acceleration methods on most visual quality metrics, which we attribute to the additional adversarial training introduced during progressive distillation. More importantly, in the 2-NFE and 1-NFE regimes, our method maintains clear advantages across the majority of evaluation metrics. Existing methods suffer from severe degradation in visual fidelity and audio–lip synchronization when pushed to the 1-NFE setting, whereas our approach remains robust.

\begin{figure*}[ht]
  \centering
  \includegraphics[width=1.0\linewidth]{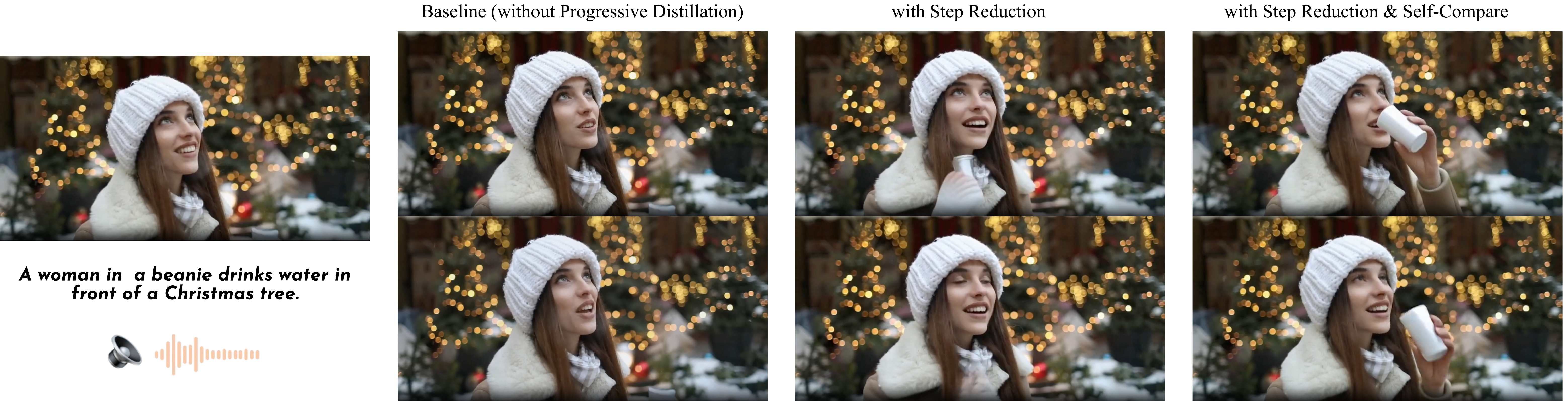}
  \caption{
    Qualitative ablation of progressive distillation under the 1-NFE setting. Step reduction enhances motion continuity, and self-compare regularization effectively prevents quality collapse, yielding stable and detailed one-step generation.
  }
  \label{fig:qualitive_ab}
\end{figure*}

Notably, even in the extreme 1-NFE configuration, our model preserves performance comparable to its 4-NFE counterpart, while achieving an additional 4× speedup. These results demonstrate the effectiveness and stability of our approach in ultra-low NFE scenarios, highlighting its suitability for real-time talking avatar generation.

\textbf{Qualitative Results.} Fig. \ref{fig:qualitive} presents a qualitative comparison between our method and representative state-of-the-art talking avatar generation approaches. Under the 4-NFE setting, our model exhibits significantly richer motion dynamics. While existing methods tend to produce limited head and hand movements, our approach generates more natural and expressive motions that respond coherently to speech variations, with notably more diverse and realistic hand gestures.

As illustrated in the right example of Fig. \ref{fig:qualitive}, when the reference image lacks explicit hand information, competing methods either fail to synthesize hands or produce visually implausible hand appearances with incorrect colors. In contrast, our method is able to hallucinate natural and high-quality hand regions consistent with the overall appearance.

Under the more challenging 1-NFE setting, our model continues to demonstrate strong robustness. Other methods suffer from severe degradation, yielding blurry facial expressions and indistinct hand structures, whereas our approach preserves visual quality comparable to that of the 4-NFE configuration. These results further highlight the effectiveness of our method in maintaining high-fidelity and expressive avatar generation under extreme inference constraints. These qualitative results indicate that our distillation framework is more stable under extreme step reduction, effectively preserving the multi-dimensional motion dynamics of the baseline video generation model. In comparison, although other methods retain basic talking functionality, they exhibit noticeable loss in expressiveness and motion diversity.

\subsection{Ablation Study}
\begin{table*}[ht]
\centering
    \caption{Ablation study of step reduction, dynamic timestep sampling, and self-compare adversarial regularization. }
     \label{Table:Ablation:main}
\begin{tabular}{ccc|ccccc}
\hline
\multicolumn{3}{c|}{Methods}                                   & \multirow{2}{*}{FID↓} & \multirow{2}{*}{FVD↓} & \multirow{2}{*}{E-FID↓} & \multirow{2}{*}{Sync-C↑} & \multirow{2}{*}{Sync-D↓} \\ \cline{1-3}
\multicolumn{1}{c}{Step Reduction} & Dynamic TS & Self-Compare &                       &                       &                         &                          &                          \\ \hline
\ding{55}           & \ding{55}          & \ding{55}            & 41.42                 & 487.32                & 3.25                    & 6.24                     & 8.91                     \\
\ding{51}                               & \ding{55}         & \ding{55}           & 35.12                 & 452.58                & 2.31                    & 7.57                     & 8.43                     \\
\ding{51}                                  & \ding{51}          & \ding{55}            &32.28        & 448.36                & 2.27                    & 7.82                     & 8.65                     \\
\ding{51}                                  & \ding{55}          & \ding{51}            & \textbf{31.41}        & 415.67                & 1.68                    & \textbf{8.63}            & 8.49                     \\
\ding{51}                                  & \ding{51}         & \ding{51}            & 31.46                 & \textbf{414.23}       & \textbf{1.61}           & 8.50                     & \textbf{8.03}            \\ \hline
\end{tabular}
\end{table*}
Table \ref{Table:Ablation:main} presents the ablation study of our proposed components. Direct one-step distillation without progressive strategies leads to consistently poor performance across all metrics, which we attribute to the instability of adversarial training under an abrupt step reduction.

Introducing step reduction significantly eases the training difficulty at each distillation stage, resulting in substantial improvements across all evaluation metrics. Further incorporating dynamic timestep sampling yields consistent, albeit moderate, gains, indicating its effectiveness in improving temporal robustness during training.

Finally, by adding self-compare adversarial regularization, the stability of adversarial training is further enhanced, leading to notable improvements in visual quality metrics. The best overall performance is achieved when all three components—step reduction, dynamic timestep sampling, and self-compare regularization—are jointly applied, demonstrating their complementary effects.

Fig.~\ref{fig:qualitive_ab} presents qualitative ablation results of our framework. When progressive adversarial distillation is removed and the model is directly distilled from 4 steps to 1 step, the student suffers from a noticeable loss of instruction-following capability. As shown in the left example in Fig.~\ref{fig:qualitive_ab}, the generated video fails to realize the “drinking water” action specified in the prompt, indicating a breakdown in high-level semantic control.

After introducing Step Reduction alone, the middle example demonstrates that the model is able to produce the target drinking motion; however, the resulting video contains substantial visual artifacts. We attribute this degradation to unstable adversarial training caused by the large distribution gap between the teacher and the single-step student.

In contrast, when all components are enabled, including progressive step reduction and self-compare adversarial regularization, the model generates videos that correctly and fully depict the drinking action with significantly improved visual quality. These results highlight the importance of progressive distillation and stabilization mechanisms for preserving instruction adherence and visual fidelity in ultra-low NFE settings.

\begin{table}[ht]
\centering
    \caption{Effect of Adversarial Loss Functions}
     \label{Table:Ablation:GAN}
\resizebox{\linewidth}{!}{
\begin{tabular}{cccccc}
\hline
Method         & \multicolumn{1}{l}{FID↓} & \multicolumn{1}{l}{FVD↓} & \multicolumn{1}{l}{E-FID↓} & \multicolumn{1}{l}{Sync-C↑} & \multicolumn{1}{l}{Sync-D↓} \\ \hline
GAN            & 33.12                    & 411.58                   & 1.80                       & 7.95                        & 8.23                        \\
Hinge         & 32.98                    & 415.91                   & 1.78                       & 8.32                        & 8.18                        \\
\textbf{R3GAN} & \textbf{31.43}           & \textbf{414.23}          & \textbf{1.61}              & \textbf{8.50}               & \textbf{8.03}               \\ \hline
\end{tabular}
}
\end{table}

We further examine whether our framework is sensitive to the specific choice of adversarial loss. We compare the standard GAN loss, hinge loss, and R3GAN under identical training and evaluation settings. As shown in Table~\ref{Table:Ablation:GAN}, all three objectives yield broadly comparable performance, indicating that our method does not rely on a particular adversarial formulation to function effectively. Among them, R3GAN consistently delivers the strongest overall results, reflecting more stable discriminator behavior and more informative gradient signals during ultra-low NFE distillation. Hinge loss offers a moderate improvement over the vanilla GAN objective, but remains less effective than R3GAN.

\begin{table}[ht]
\centering
    \caption{Effect of Self-Compare Regularization Weight }
     \label{Table:Ablation:lambda}
\begin{tabular}{cccccc}

\hline
$\lambda$ & \multicolumn{1}{l}{FID↓} & \multicolumn{1}{l}{FVD↓} & \multicolumn{1}{l}{E-FID↓} & \multicolumn{1}{l}{Sync-C↑} & \multicolumn{1}{l}{Sync-D↓} \\ \hline
10          & 32.57                    & 448.51                   & 2.31                       & 7.76                        & 8.63                        \\
30          & \textbf{31.37}                   & 424.37                   & 1.85                       & 8.06                        & 8.47                        \\
\textbf{50} & 31.46          & \textbf{414.23}          & \textbf{1.61}              & 8.50               & \textbf{8.03}               \\
70          & 31.87                    & 416.76                   & 1.76                       & \textbf{8.72}                        & 8.21                        \\
90          & 31.75                    & 416.41                   & 1.74                       & 8.23                        & 8.25                        \\ \hline
\end{tabular}
\end{table}

We then analyze the role of self-compare regularization, which is central to stabilizing progressive adversarial distillation. Table~\ref{Table:Ablation:lambda} reports results under varying values of the weighting coefficient $\lambda$. The performance exhibits a clear U-shaped trend: overly weak regularization fails to sufficiently constrain adversarial optimization, while excessive weighting over-anchors the student to the four-step reference and limits adaptation. A moderate setting ($\lambda=50$) strikes the best balance, leading to consistently improved perceptual quality and synchronization. This observation supports our design motivation that self-compare regularization acts as an effective intermediate guidance mechanism, easing optimization by bridging real data and single-step student outputs.

%% file: sec/5_conclusion.tex
\section{Conclusion}
We propose TurboTalk, a progressive distillation framework that enables stable and high-quality audio-driven talking avatar generation under ultra-low NFE, down to a single denoising step. By combining distribution matching distillation with progressive adversarial distillation, dynamic timestep sampling, and self-compare regularization, TurboTalk effectively mitigates the instability and quality collapse commonly observed in extreme few-step settings. Extensive experiments show that our method achieves up to 120× inference speedup over strong diffusion-based baselines while preserving expressive facial dynamics and accurate audio–lip synchronization.
These results highlight TurboTalk as a practical solution for real-time digital human generation and shed light on robust distillation of diffusion models in ultra-low-step regimes.